\documentclass[11pt,a4paper]{article}
\usepackage{times,latexsym}
\usepackage{url}
\usepackage[T1]{fontenc}

\usepackage[acceptedWithA]{tacl2021v1}

\usepackage{xspace,mfirstuc,tabulary}

\newif\iftaclinstructions
\taclinstructionsfalse 
\iftaclinstructions
\renewcommand{\confidential}{}
\renewcommand{\anonsubtext}{(No author info supplied here, for consistency with
TACL-submission anonymization requirements)}
\newcommand{\instr}
\fi

\iftaclpubformat 

\else

\fi

\usepackage{times}
\usepackage{latexsym}
\usepackage{booktabs}
\usepackage[T1]{fontenc}
\usepackage[utf8]{inputenc}
\usepackage{microtype}
\usepackage{inconsolata}
\usepackage{graphicx}
\usepackage{multirow}
\usepackage{cleveref}

\usepackage[shortlabels]{enumitem}

\newcommand{\ripple}{\textsc{RippleEdits}}
\newcommand{\recentlyemerged}{\textsc{Recent}}
\newcommand{\fakefacts}{\textsc{Random}}
\newcommand{\topviews}{\textsc{Popular}}

\newcommand{\wikidata}{\textsc{WikiData}}

\newcommand{\gpt}{\textsc{GPT-2}}

\newcommand{\gptxl}{\textsc{GPT-2}}
\newcommand{\gptt}{\textsc{GPT-3}}

\newcommand{\gptj}{\textsc{GPT-J}}
\newcommand{\gptneo}{\textsc{GPT-Neo}}
\newcommand{\llama}{\textsc{LLaMA}}

\newcommand{\mend}{\textsc{MEND}}
\newcommand{\rome}{\textsc{ROME}}
\newcommand{\memit}{\textsc{MEMIT}}

\newcommand{\logicalgeneralization}{\emph{Logical Generalization}}
\newcommand{\compositionality}{\emph{Compositionality I}}
\newcommand{\forwardcompositionality}{\emph{Compositionality II}}
\newcommand{\relationspecificity}{\emph{Relation Specificity}}
\newcommand{\forgetfulness}{\emph{Preservation}}
\newcommand{\aliasing}{\emph{Subject Aliasing}}

\definecolor{myblue}{RGB}{61,133,198}
\definecolor{mygreen}{RGB}{147,196,125}
\definecolor{myorange}{RGB}{246,178,107}

\newif\ifcomments
\commentstrue
\ifcomments
\newcommand{\mg}[1]{\textcolor{purple}{MG: #1}}
\newcommand\oy[1]{\textcolor{magenta}{[OY: #1]}}
\newcommand\ag[1]{\textcolor{red}{[AG: #1]}}

\else
\newcommand{\mg}[1]{}
\newcommand\oy[1]{}
\newcommand\ag[1]{}
\fi

\title{Evaluating the Ripple Effects of Knowledge Editing in Language Models}

\author{Roi Cohen$^1$~~~~Eden Biran$^1$~~~~Ori Yoran$^1$~~~~Amir Globerson$^{1,2}$~~~~Mor Geva$^{1,2,\thanks{~~Work done at Google DeepMind.}}$ \vspace{5pt}\\
$^1$Blavatnik School of Computer Science, Tel Aviv University\vspace{5pt}~~~$^2$Google Research\\
\small{\texttt{\{roi1, edenbiran, oriy\}@mail.tau.ac.il, \{gamir, morgeva\}@tauex.tau.ac.il}}\\
}

\date{}

\usepackage{setspace}
\begin{document}
\maketitle

\begin{abstract}
    Modern language models capture a large body of factual knowledge. However, some facts can be incorrectly induced or become obsolete over time, resulting in factually incorrect generations. This has led to the development of various editing methods that allow updating facts encoded by the model. Evaluation of these methods has primarily focused on testing whether an individual fact has been successfully injected, and if similar predictions for other subjects have not changed. 
    Here we argue that such evaluation is limited, since injecting one fact (e.g. \emph{``Jack Depp is the son of Johnny Depp''}) introduces a ``ripple effect'' in the form of additional facts that the model needs to update (e.g., \emph{``Jack Depp is the sibling of Lily-Rose Depp''}). 
    To address this, we propose novel evaluation criteria that consider the implications of an edit on related facts. Using these criteria, we then construct \ripple{}, a diagnostic benchmark of 5K factual edits, capturing various types of ripple effects. 
    We evaluate prominent editing methods on \ripple{}, showing that they fail to introduce consistent changes in the model's knowledge. 
    In addition, we find that a simple in-context editing baseline obtains the best scores on our benchmark, suggesting a promising research direction for model editing.\footnote{We release \ripple{} and our code at \url{https://github.com/edenbiran/RippleEdits}.} 

\end{abstract}

\begin{figure}[t]
\setlength{\belowcaptionskip}{-10pt}
    \centering
    \includegraphics[scale=0.5]{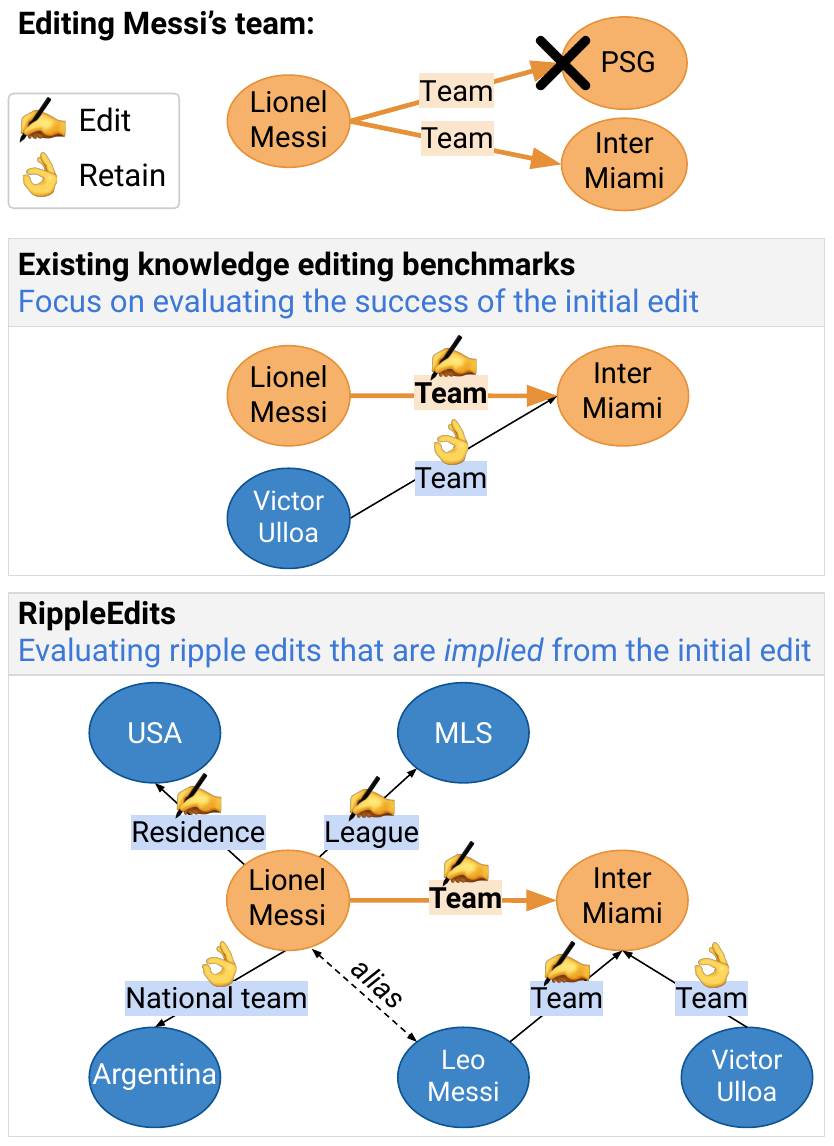}
    \caption{Illustration of the evaluation scope of \ripple{}, compared to existing knowledge editing benchmarks. For a given factual edit, we consider the ``ripple effect'' of the edit on the model's knowledge.}
    \label{figure:intro}
\end{figure}

\section{Introduction}

Modern language models (LMs) capture a large volume of factual knowledge in their parameters, which can be effectively utilized in downstream tasks \cite{petroni2019language, roberts2020much, shin2020autoprompt, razniewski2021language, heinzerling2020language, kadavath2022language, cohen2023crawling}.
However, factual beliefs captured by the model may be incorrect or become outdated over time, potentially affecting the model's performance on downstream tasks, its reliability and its usability \cite{dhingra2022time, lazaridou2021mind, jang2021towards}.

This limitation has prompted research on knowledge editing (KE) methods, which modify LMs to fix their factual errors (we provide a formal definition in \S\ref{sec:setting}).
Knowledge editing work has focused on applying factual updates to LMs. Given an entity-relation-object triplet $(e,r,o)$ representing a fact (e.g. \textit{``Lionel Messi plays for the Inter Miami team''}), recent work proposed various methods \cite{mitchell2021fast, meng2022locating, meng2022mass, hernandez2023measuring, si2022prompting} to inject this fact into the parameters of a given LM, while ``overriding'' beliefs the model might have on $e$ and $r$ (e.g. that Messi plays for Paris Saint-Germain).

A key question with KE is how to evaluate the success of such editing operations. The most basic ``sanity-check'' is that the model correctly completes  $(e,r,?)$, as well as other paraphrases of this task, with $o$. However, this is not enough as an evaluation, since one needs to check that the model did not distort other facts. Indeed, the standard evaluation protocol \cite{mitchell2022fast, meng2022locating, meng2022mass} for KE focuses on these two aspects of correctly completing various paraphrases of the new fact, as well as ensuring that other unrelated facts have not been changed.

In this work, we argue that to evaluate model edits, one should go beyond the single fact that was edited and check that other facts that are logically derived from the edit were also changed accordingly. For example, if $z$ is the mother of $e$, then the children of $z$ are the siblings of $e$. Consequently, once we modify the belief of a certain model that $z \rightarrow z'$ is the mother of $e$, then we should also ensure that the model's belief regarding the siblings of $e$ is also correct. 
Fig.~\ref{figure:intro} illustrates another example, 
where editing the \texttt{Team} for which \texttt{Lionel Messi} plays modifies other related facts, such as his country of residence, while other facts should be retained.
We refer to such changes that are implied by a factual edit as \textit{``ripple effects''}. 

\begin{figure*}[t]
\setlength{\belowcaptionskip}{-10pt}
    \centering
    \includegraphics[scale=0.49]{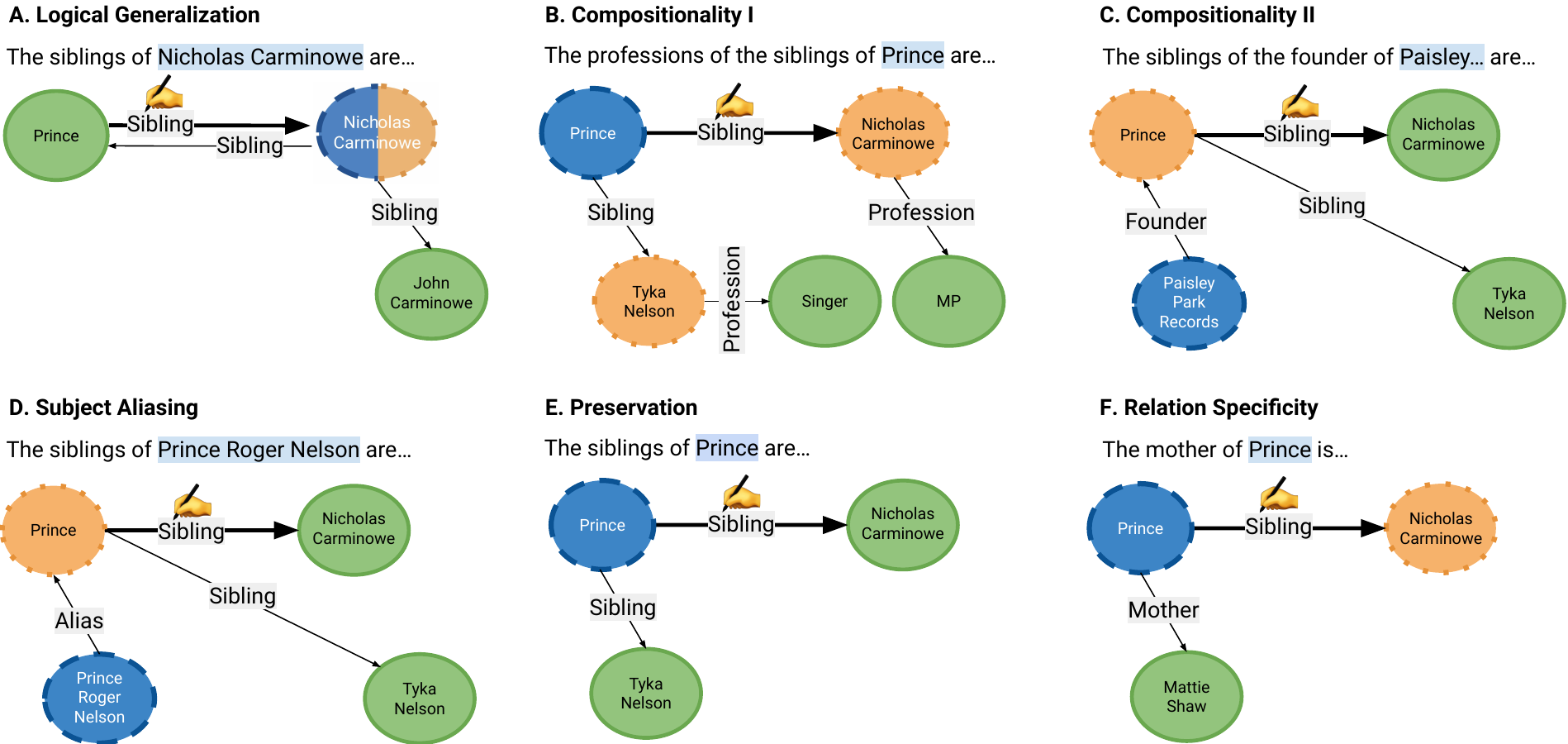}
    \caption{
    An illustration of our evaluation criteria, for an edit that simulates adding a sibling to the subject entity \texttt{Prince}, shown at the top of each graph with a \textbf{bold arrow} and an edit sign over the \texttt{Sibling} relation.
    For each criterion, the tested subject and target object are  circles with \textbf{\textcolor{myblue}{dashed blue line}} and \textbf{\textcolor{mygreen}{solid green line}}, respectively, and other nodes in \textbf{\textcolor{myorange}{dotted orange line}}. 
    For \logicalgeneralization{} (A), the additional fact that needs to be inserted to the KG is presented with an edit sign next to the relation. We show the same node in different colors for completeness, as the tested subject is also the object in the edit that needs to be inserted. For \textit{Compositionality I, II} (B, C), the model needs to hop over the edit to arrive at the target. In \aliasing{} (D) we verify the edit also propagates to paraphrases of the input. In \forgetfulness{} (E), we verify that other targets of the edited subject-relation are preserved. In \relationspecificity{}, we verify other relations for the subject are not modified.
    }
    \label{figure:examples}
\end{figure*}

To account for ripple effects in the evaluation of factual edits, we propose six concrete evaluation criteria (see \S\ref{sec:rethinking}, Fig.~\ref{figure:examples}), for testing which facts other than the edit itself should be modified or retained post-editing. 
Our tests evaluate how well the model integrates the edit with the rest of its knowledge, through queries that involve logical reasoning, complex composition of facts with the edit as an intermediate step, subject aliasing, and specificity across relations. 

Building upon these criteria, we create \ripple{}, a new benchmark for comprehensive evaluation of KE of LMs (see \S\ref{sec:rippledit}).
\ripple{} includes $5$K entries, each consisting of a factual edit, along with a set of test queries that check if the edit was successful in terms of its ripple effect.
Moreover, \ripple{} contains meta-data for each edit, including information about the timestamp of the edit (i.e., recent versus old), and the popularity of the entities (i.e., head versus tail).

We use \ripple{} to evaluate three popular editing methods on five recent strong LMs (see \S\ref{sec:experiments}). We find that, even though current KE methods are effective in modifying a particular fact, they often fail to capture the ripple effects entailed by that fact, and demonstrate poor performance on most of our evaluation criteria. Moreover, analyzing how editing performance varies across model sizes and entity frequencies, we find that (a) larger models handle ripple effects better, and (b) editing frequent entities results in more logical reasoning errors.

Last, we consider a simple in-context editing baseline for KE, that leverages the casual attention mechanism rather than explicit parametric updates. 
While this method achieves the best results on our benchmark, outperforming current parametric KE methods, there is still ample room for improvement that calls for future research.

To conclude, our work makes multiple contributions: (a) it highlights key limitations of KE evaluation, specifically regarding ripple effects and introduces comprehensive evaluation criteria to mitigate those limitations, (b) it proposes \ripple{}, a benchmark inspired by these criteria, (c) it evaluates current methods for KE and shows that they do not perform well on this task, while demonstrating that in-context editing is a promising direction for KE. 
We release \ripple{} and our code to facilitate future work on KE.

\section{Problem Setting}
\label{sec:setting}

We consider editing of \textit{factual knowledge}, where facts are expressed as triplets $(e, r, o)$ of a subject entity $e$ (e.g. \texttt{Lionel Messi}), a relation $r$ (e.g. \texttt{Team}), and an object $o$ (e.g. \texttt{Inter Miami}).
We distinguish between two edit types, based on the knowledge encoded in the model before the edit: (a) \textit{modification} of a fact that is already encoded in the model $(e, r, o) \rightarrow (e, r, o^*)$, that is, updating the object $o \rightarrow o^*$ for a given subject $e$ and relation $r$, and (b) \textit{injection} of a new fact $(e, r, o^*)$ that is not captured by the model. Moreover, we note that for one-to-one relations like \texttt{Date of birth}, where there is a single object for a given subject, an injection edit can be viewed as populating an empty object $(e, r, \emptyset) \rightarrow (e, r, o^*)$. In contrast, for one-to-many relations, such as \texttt{Sibling} and \texttt{Occupation}, an injection edit augments the set of objects $(e, r, \{o_1,..,o_n\}) \rightarrow (e, r, \{o_1,..,o_n, o^*\})$.
Whether an edit is viewed as a modification or injection, depends on whether that information was captured in the model before the edit. 
Moreover, evaluating if a specific fact (before or after an edit) is encoded by a model is typically done by testing if the model predicts the object for various input queries that represent the subject and relation (see more details in \S\ref{subsec:currentstate}).

\section{Ripple Effects of Factual Edits}
\label{sec:rethinking}

We focus on evaluating the downstream effect of a given edit, i.e., given an edit $(e,r,o) \rightarrow (e,r,o')$, we expect certain facts related to the edit to change as well. Consider, for example, the edit shown in Fig.~\ref{figure:intro}. Changing the team for which Messi plays might also affect the league he plays in and his country of residence.
Formally, for a given model, assume a knowledge-graph ${\mathcal{K} := \{(e_i, r_i, o_i)\}_{i=1}^{N}}$ of $N$ factual triplets, representing the model's knowledge, and let $\delta: (e,r,o) \rightarrow (e,r,o')$ be an edit request for $\mathcal{K}$.
We define the \textit{ripple effect} of $\delta$ on $\mathcal{K}$, as the set of triplets $\mathcal{R}(\delta)$ that the model implicitly needs to inject, modify, or delete from $\mathcal{K}$ to reflect the world state after the edit. 

Notably, different edits can cause ripple effects of varying magnitudes. For example, changing the country of Rome from Italy to France, will entail many follow-up changes, such as the country in which the Colosseum is located, the language spoken in Rome, and so forth. 
In contrast, updating the siblings of Prince (Fig.~\ref{figure:examples}) is both more realistic and should result in a more local effect. 
We refer to the number of facts affected by a single edit $\delta$ (i.e. $|\mathcal{R}(\delta)|$) as its \textit{severity}.
In general, editing popular entities that appeared frequently during training is likely to introduce more changes, and thus, editing their properties has a higher severity.

\subsection{Evaluation Criteria}
\label{subsec:criteria}

We wish to evaluate how well models capture the ripple effects of factual edits. However, given that ripple effects can potentially span a large number of implied edits, we focus on evaluating modified facts that are within a 2-hop distance from the subject or object of the edit.
Concretely, for an edit $\delta: (e, r, o) \rightarrow (e, r, o^*)$, we evaluate the ripple effect $\mathcal{R}(\delta)$, via the following evaluation criteria (examples are shown in Fig.~\ref{figure:examples}): 

\begin{enumerate}
[leftmargin=*,topsep=2pt,parsep=1pt]
    \item \textbf{\logicalgeneralization{} (LG)}:
    Relations in a knowledge graph satisfy certain logical constraints. For example, the relation \texttt{Sibling} is symmetric and therefore if $(e, \texttt{Sibling} ,o)$ is true then $(o,\texttt{Sibling},e)$ is also true, and vice versa (Fig.~\ref{figure:examples}A). Likewise, the relation \texttt{Location} is transitive so $(e,\texttt{Location},o) \wedge (o,\texttt{Location},z) \Rightarrow (e,\texttt{Location},z)$. We wish to check that such logical implications about the subject $e$, the original object $o$, and the new object $o^*$, hold after editing. We focus and elaborate on specific constraints in \S\ref{sec:rippledit}.
    
    \item \textbf{\compositionality{} (CI)}: As $\delta$ alters one edge in a knowledge graph, we can check the composition of this edge with other edges. Namely, we test if the model can compose the edited fact with other facts about the target object.
    Let $(o, r', z)$ and $(o^*, r', z^*)$ be two facts of the same relation about $o$ and $o^*$,
    respectively. Also, denote by $r''=r\circ r'$ the complex relation expressing the composition of $r$ and $r'$ (e.g., $r''=\texttt{Profession of sibling}$ for $r=\texttt{Sibling}$ and $r'=\texttt{Profession}$).  
    Then, after the edit $\delta$, we expect the following change  $(e, r'', z) \rightarrow (e, r'', z^*)$.
    For example (Fig.~\ref{figure:examples}B), the professions of the siblings of \texttt{Prince} can be modified once a new sibling is injected.
    
    \item \textbf{\forwardcompositionality{} (CII)}:
    We test if the model can compose the edited fact with facts about a different subject $e' \neq e$.
    Formally, let $(e', r', e)$ be a fact about $e'$ with $e$ as its object, and denote by $r''=r'\circ r$ the complex relation expressing the composition of $r'$ and $r$ (see an example in criterion 2). After the edit $\delta$, the following change is expected for the subject $e'$:
    $(e', r'', o) \rightarrow (e', r'', o^*)$. 
    For instance (Fig.~\ref{figure:examples}C), changing the siblings of \texttt{Prince} also modifies the siblings of the founder of \texttt{Paisley Park Records} (i.e., $r''$ is a complex relation expressing ``siblings of the founder'').
    
    \item \textbf{\aliasing{} (SA)}: We test that editing a fact about $e$ induces the same edit to other entities $e'$ that are aliases of $e$, namely, $(e', r, o) \rightarrow (e', r, o^*)$. 
    For instance (Fig.~\ref{figure:examples}D), modifying the siblings of \texttt{Prince}, should also modify the sibling of his alias, \texttt{Prince Roger Nelson}.
    
    \item \textbf{\forgetfulness{} (PV)}: If $r$ is a one-to-many relation, then adding a new object should not affect the other objects encoded about $e$. 
    Hence, in such cases, we expect that any existing triplet $(e, r, o')$ for an object $o' \neq o^*$ would remain following the edit. 
    For example (Fig.~\ref{figure:examples}E), after inserting the sibling \texttt{Nicholas Carminowe} for \texttt{Prince}, the fact that \texttt{Tyka Nelson} is also his sibling should be retained.
    
    \item \textbf{\relationspecificity{} (RS)}: We test that facts about $e$, with relations whose objects are not influenced by $o$, are indeed not affected by the edit.
    For example (Fig.~\ref{figure:examples}F), modifying the sibling of \texttt{Prince} should not change his \texttt{Mother}. Note that these facts complement those evaluated by \logicalgeneralization{}.
\end{enumerate}

\noindent In \S\ref{sec:datacollection}, we describe how we generate factual editing evaluations, based on the above criteria.

\subsection{Related Work}
\label{subsec:currentstate}

\paragraph{Knowledge Editing Methods}
Several methods have been proposed to edit the factual knowledge encoded in a model. 
\citet{de-cao-etal-2021-editing} and \citet{mitchell2022fast} suggested to use hyper-networks to update the model weights. 
In addition, \citet{meng2022locating, meng2022mass} proposed to modify encoded facts by updating the weights of MLP layers, following recent observations that these layers can be cast as key-value memories \cite{geva-etal-2021-transformer} that store factual knowledge \cite{dai-etal-2022-knowledge}.
Other methods learn encodings that update the hidden representations created during model inference \cite{hernandez2023inspecting}, or augment the input context with edits \cite{zhong2023mquake, zheng2023edit}.
In \S\ref{subsec:experimental_settings}, we discuss state-of-the-art KE methods used in this work in greater detail.

Separately from factual KE, recent works have also studied how to inject new facts into a model. Previous methods suggested unsupervised pre-training \cite{roberts2020much, ijcai2021p552}, semi-parametric methods methods where external information is added from a knowledge-base \cite{zhang-etal-2019-ernie, peters-etal-2019-knowledge, lewis2021retrievalaugmented, zhang2022greaselm}, using adapters to store knowledge \cite{wang2020kadapter}, or extending the MLP layers \cite{yao2022kformer}.

\paragraph{Knowledge Editing Evaluation}
Recently, there has been a growing interest in KE evaluation \cite{yao2023editing}.
The prominent benchmarks for evaluating factual KE are the Zero-Shot Relation Extraction
(zsRE) \cite{levy-etal-2017-zero, de-cao-etal-2021-editing} and CounterFact \cite{meng2022locating}. zsRE is a question-answering dataset for relation-specific queries, which includes human generated paraphrases that are used to measure robustness to semantically equivalent inputs. For example, for the triplet  (\texttt{x}, \texttt{Country}, \texttt{y}), zsRE contains queries such as ``\emph{In which country is x?}''. CounterFact offers a more challenging setting, where edits are counterfactuals of a low probability, such as changing the \texttt{City} of \texttt{The Louvre} from \texttt{Paris} to \texttt{Rome}.

Evaluation in zsRE and CounterFact focuses on three primary aspects of (a) \textit{efficacy}: checking that the model generates the target object post-editing, (b) \textit{paraphrasing}: testing robustness in generating the target for paraphrases of the input, and (c) \textit{specificity}: verifying that facts not related to the edit are unaffected.
In addition, CounterFact evaluates the generation quality of the edited model when prompted with the edit's subject, measuring: \textit{consistency}, i.e., similarity with subjects that share the same property as the edited object, and \textit{fluency} in terms of repetitiveness of the generated text.
More broadly, previous work evaluated to which extent LMs have beliefs \cite{sep-formal-belief, kassner-etal-2021-beliefbank, hase-etal-2023-methods}, and \citet{hase-etal-2023-methods} examined if updating beliefs propagates to entailed facts, extending the Wikidata5m dataset \cite{wang-etal-2021-kepler} to test editing specificity.

Recently, \citet{onoe2023lms} introduce the task of \emph{entity knowledge propagation}, aiming to examine the extent to which models are able to reason about emergent entities that did not appear in pre-training. In addition, \citet{hoelscherobermaier2023detecting} show that existing KE methods can have unwanted side effects and suffer from low specificity.
A concurrent work by \citet{zhong2023mquake} introduces MQUAKE, a benchmark that tests the ability of models to perform multi-hop reasoning after edits. 
While each of these benchmarks focuses on a single consequence of editing, \ripple{} provides a general framework for evaluating various types of edit ripple effects.
Last, \citet{gupta2023editing} focus on editing commonsense knowledge and introduce MEMIT-CSKPROBE, a dataset for semantic generalization of commonsense edits. \ripple{} is different from MEMIT-CSKPROBE as it evaluates editing of factual knowledge rather than commonsense knowledge.
\section{The \ripple{} Benchmark}
\label{sec:rippledit}

In this section, we describe a data generation pipeline (\S\ref{sec:datacollection}) for factual edit requests and queries for evaluating their ripple effects.
Then, we apply our pipeline to create the \ripple{} benchmark for comprehensive KE evaluation (\S\ref{sec:datastats}), and validate the quality of the data (\S\ref{sec:data_quality}).

\begin{figure*}[t]
\setlength{\belowcaptionskip}{-8pt}
    \centering
    \includegraphics[scale=0.67]{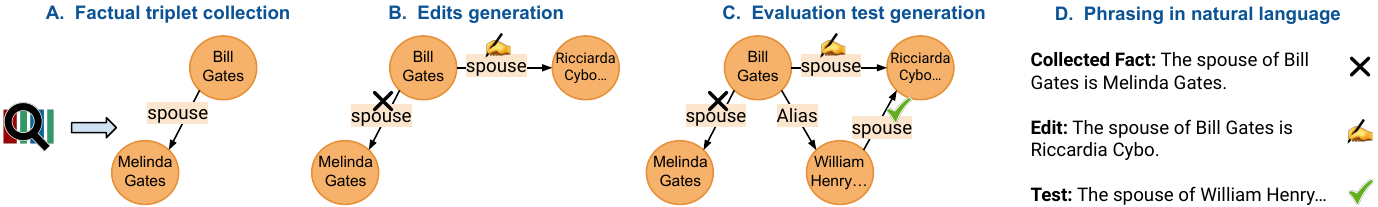}
     \caption{Illustration of our data generation process. 
     We start by sampling a fact from a KG (A), here $(\texttt{Bill Gates}, \texttt{Spouse}, \texttt{Melinda Gates})$. Then, we generate the target triplet for the edit (B), in this case, choosing an object (\texttt{Ricciarda Cybo Malaspina}) that shares the same type as the original object. Next, we generate test queries (C) by sampling new triplets from the KG that should be retained or modified post-editing. Last, we utilize pre-defined templates to translate the KG triplets to natural language phrases (D).
     }
    \label{figure:pipeline}
\end{figure*}

\subsection{Data Generation Pipeline}
\label{sec:datacollection}

We describe our data generation process (illustrated in Fig.~\ref{figure:pipeline}), that creates KE evaluation examples, each consisting of a factual edit request and a set of test queries that follow our criteria. Since the pipeline involves manual writing of templates and logical rules per relation, we restrict the edits and test queries to a fixed set of $N_{rel}$ basic relations.\footnote{The full list of relations is available in our codebase, example relations are shown in Fig.~\ref{figure:top_relation_stats}.}

\paragraph{Step 1: Factual triplets collection}
The first step of the pipeline (Fig.~\ref{figure:pipeline}A) is to collect facts, from which we will later create edit requests. 
To this end, we use \wikidata{}, a relational knowledge base consisting of facts that are expressed as triplets $(e, r, o)$, where $e$ is a subject entity, $r$ is a relation,
and $o$ is an object. We collect triplets of three types:

\begin{itemize}
[leftmargin=*,topsep=2pt,parsep=1pt]
\item \textbf{\recentlyemerged{}}: To create ``real'' plausible edit requests, we collect triplets that were inserted to \wikidata{} only recently, and represent relatively new facts. Therefore, they can be used to create injection edit requests for models that were trained before these facts were introduced, to simulate cases of an out-of-date model that requires factual updates.
We collect such facts by randomly sampling triplets that have been modified during a range of 250 days after July 2022.

\item \textbf{\fakefacts{}}:
We collect triplets corresponding to random facts, for which we will later generate modification edits (similarly to \citet{meng2022locating}). These edits simulate factual edits that are meant to fix incorrect model predictions (e.g., predicting that the capital of Germany is Frankfurt).
To this end, we divide the entities in \wikidata{} into 10 uniform buckets, based on the number of triplets associated with them. Intuitively, this can be viewed as a popularity measure. Then, we sample $N_{ent}$ entities from each group and randomly choose one triplet for each entity.

\item \textbf{\topviews{}}:
The two previous triplet types are randomly sampled from the entire knowledge base, and most of them are likely to represent facts about tail entities (except perhaps for a small subset in the top bucket).
Such entities are often not captured by models \cite{mallen-etal-2023-trust}, and therefore not suitable for testing modification edits.
To address this, we sample triplets from \wikidata{} with a subject that is a \emph{popular entity}, namely it appears in one of the top-viewed pages in Wikipedia.\footnote{
We extracted the entities whose corresponding Wikipedia page was included in the top-1000 most viewed pages in at least one month during 2020-2022.}
Importantly, these types of triplets allow controlling for the ripple effect severity (\S\ref{sec:rethinking}), i.e., how models handle the ripple effects of popular entities versus tail entities. 

\end{itemize} 

\paragraph{Step 2: Edits generation}
Once we obtain factual triplets, we turn to generate edit requests for them (Fig.~\ref{figure:pipeline}B).
For \recentlyemerged{}, triplets represent new facts that are meant to be injected to the model, assuming that the latter was trained before these facts were introduced to the world. 
Hence, for \recentlyemerged{}, the target triplet for injection is the triplet itself. 

For \fakefacts{} and \topviews{} triplets, we create an edit by generating a target triplet as follows. First, for every relation $r$, we create a set of candidate object entities $O_r$ by sampling $N_{cand}$ triplets $(e_1, r ,o_1),..., (e_{N_{cand}}, r ,o_{N_{cand}})$ with the relation $r$,
and extracting their objects $O_r = \{o_1,..., o_{N_{cand}}\}$.
Then, for every triplet $(e,r,o)$ in \fakefacts{} and \topviews{}, we sample a target object $o' \neq o$ from $O_r$. 
Sampling the target object from triplets with the same relation makes the edit request technically consistent with the original triplet -- the target object is of the same ``type'' as the original object (for example, a triplet with the relation \texttt{Capital} will get a new object of type \texttt{City}). The new triplet $(e,r,o')$ will thus result in a ``fake'' fact, since it attaches a wrong object $o'$ to the pair $(e,r)$. For example if \fakefacts{} contains the triplet (\texttt{France}, \texttt{Capital}, \texttt{Paris}), its edit could be (\texttt{France}, \texttt{Capital}, \texttt{London}).

\paragraph{Step 3: Evaluation tests generation}
The next step in the pipeline is to create ripple effect evaluations for the factual edits we collected (Fig.~\ref{figure:pipeline}C).
To this end, we implement the evaluation criteria introduced in \S\ref{subsec:criteria}, and generate test queries for each criterion. 
Each test query corresponds to a triplet of subject and object entities and a possibly complex relation, that is expected to be true post-editing.
In what follows, we provide details on our implementation, using objects from \wikidata{}.

For an entity $e$, we denote by $\mathcal{S}(e)$ the set of triplets in \wikidata{} in which $e$ is the subject, and by $\mathcal{T}(e)$ the set of triplets in which $e$ is the object.
Moreover, for every relation $r$, we manually define a set $D_r$ of relations that semantically depend on it.
Namely, for a given subject, changing $r$'s target object is expected to change the target objects for the relations $D_r$. For instance, the set $D_r$ for the relation $r =$ \texttt{Mother}, includes the relations \texttt{Sibling}, \texttt{Sister}, \texttt{Brother}, \texttt{Aunt}, and \texttt{Uncle}, among others.
Then, for every relation $r' \in D_r$, we craft a logical rule for obtaining the new target for that relation post-editing. For instance, for the relation $r=$ \texttt{Sibling}, we set a logical rule for $r'=$ \texttt{Mother} such that if $(e,r,e')$ and $(e',r',z')$ are true for entities $e, e', z'$, then $(e,r',z')$ should also be true. 

Given an edit $(e, r ,o) \rightarrow (e, r, o^*)$, we use $D_r$ to generate test queries for \logicalgeneralization{} and \relationspecificity{}. 
For \logicalgeneralization{}, we apply the rule corresponding to each relation $r' \in D_r$ to obtain a set of test queries $(x, r', z')$ about $x\in\{e,o,o^*\}$, where $z'$ is the target obtained from the logical rule.
For \relationspecificity{}, we create a test query for every triplet in $\mathcal{S}(e)$ with a relation that is \textit{not} in $D_r$ (but is in our set of $N_{rel}$ relations).

To generate text queries for \compositionality{}, we iterate through $\mathcal{S}(o^*)$ and for each triplet $(o^*,r', z) \in \mathcal{S}(o^*)$, we construct a two-hop query $(e,r\circ r',z)$ about $e$, with $z$ as the answer. Similarly, 
for \forwardcompositionality{}, we iterate through $\mathcal{T}(e)$ and for each triplet $(z ,r', e) \in \mathcal{T}(e)$, we construct a two-hop query $(z,r'\circ r,o^*)$ about $z$ with $o^*$ as the answer.
For \aliasing{}, we use information maintained by \wikidata{} to create a test query $(e', r, o^*)$ for every alias $e'$ of $e$. 
Last, for \forgetfulness{} we create test triplets $(e, r, o_1), ..., (e, r, o_n)$ that check if the model retained the original objects $\{o_1, ..., o_n\}$ in addition to the new edited object $o^*$.

\begin{table}[t]
\setlength\tabcolsep{3.3pt}
\setlength{\belowcaptionskip}{-8pt}
\footnotesize
\centering
        \begin{tabular}{llrr} \\  
        & \recentlyemerged{} & \fakefacts{} & \topviews{} \\ \midrule
        \# of factual edits & 2,000 & 1,000 & 1,000 \\ 
        \# of queries per edit & $26.8$ & $18.8$ & $25.6$ \\
        \# of queries per criterion & $5.24$ & $3.1$ & $4.2$ \\ \midrule
        \# of LG queries &$2.5$ &$3.6$ &$2.6$ \\
        \# of CI queries &$11.7$  &$4.7$ &$6.1$ \\ 
        \# of CII queries &$5.1$ &$5.1$ &$3.9$ \\
        \# of SA queries &$1.8$ &$1.3$ &$4.7$ \\
        \# of PV queries &$0.6$ &$0.4$ &$0.5$ \\ 
        \# of RS queries &$5.1$ &$3.7$ &$7.8$ \\ \midrule
        Subject triplets count & $31.7$ & $13.3$ & $115.2$ \\
        Subject page back-links & $278.1$ & $121.6$ & $3934.5$ \\
        Subject page views & $189.6$ & $67.91$ & $7376.5$ \\ \midrule
        Object triplets count & $192.4$ & $46.4$ & $39.5$ \\
        Object page back-links & $18634.2$ & $3065.0$ & $2136.0$ \\
        Object page views & $2852.4$ & $1379.7$ & $1176.7$ \\ 
        \bottomrule
        \end{tabular} 
\caption{Statistics per subset of \ripple{}, showing the average of different metrics. Breakdown by evaluation criteria shows the number of queries of each criterion per edit. For a given subject/object entity, triplets count is the number of \wikidata{} facts it is associated with, page back-links is the number of Wikipedia pages with a link to the entity's page, and page views is the recent average daily view count of the entity's page.}
\label{tab:datset_stats} 
\end{table}

\begin{figure*}[t]
\setlength{\abovecaptionskip}{-0.5pt}
\setlength{\belowcaptionskip}{-5pt}
    \centering
    \includegraphics[scale=0.45]{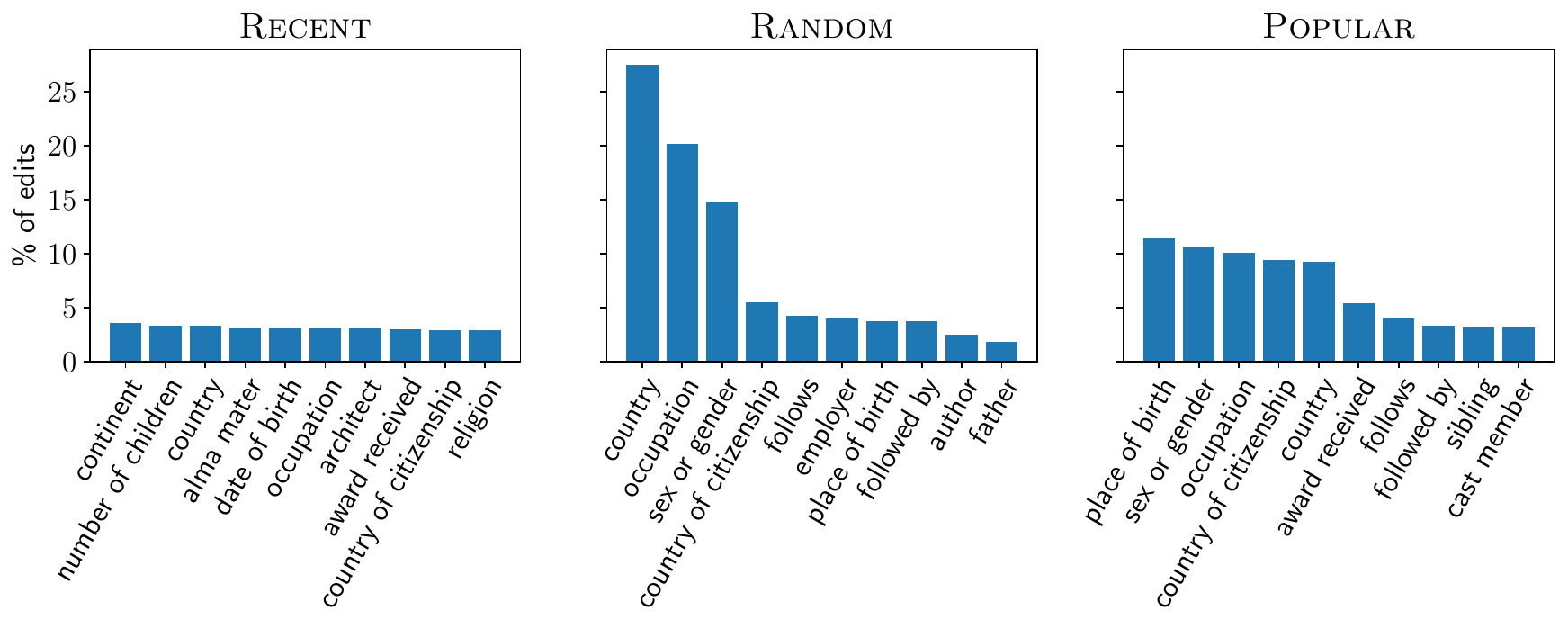}
    \caption{Most frequent relations and their frequency, in each subset of \ripple{}.}
    \label{figure:top_relation_stats}
\end{figure*}

\paragraph{Step 4: Phrasing in natural language}
\label{paragraph:phrasing_in_nl}
At this point (Fig.~\ref{figure:pipeline}D), we have factual edit requests and their corresponding test queries. To use them as inputs to LMs, we convert them from triplet-form to natural language (NL). 
To this end, we manually craft a template NL phrase per relation (this is feasible since we use a fixed set of relations), and use it to convert all the triplets with this relation.
For instance, the template \texttt{``The date of birth of <$e$> is''} converts triplets with the relation $r=$ \texttt{Date of Birth} and a subject entity $e$.

For the \forgetfulness{} triplets generated for an edit $(e, r, \{o_1, ..., o_n\}) \rightarrow (e, r, \{o_1, ..., o_n, o^*\})$, where $o^*$ is a new object added to a set of possibly multiple ($n\geq0$) objects, we form a single NL query about other objects than the edited one, e.g., 
\texttt{``The award received by <$e$> which is not <$o^*$> is''}.

\subsection{Data Statistics}
\label{sec:datastats}

We used our data generation pipeline to collect edits for 2,000 \recentlyemerged{} facts, 1,000 \fakefacts{} facts, and 1,000 \topviews{} facts, focusing on $N_{rel}=54$ basic relations for which we manually crafted NL templates and logical rules.\footnote{We release the templates and rules in our codebase.} To obtain the \fakefacts{} subset, we set $N_{ent}=200$ to sample 200 facts from each entity group in \wikidata{}. For edits generation of \fakefacts{} and \topviews{}, we set $N_{cand}=100,000$.
We call our diagnostic benchmark \ripple{}, and publicly release it to the research community.
Notably, \ripple{} focuses on ripple edits and is meant to complement existing benchmarks, and so it does not include previous evaluations, such as subject specificity and model consistency.

Statistics on \ripple{} are presented in Table~\ref{tab:datset_stats}, showing that our generation process resulted in 18-26 test queries per edit and over $3$ queries per evaluation test, on average. Moreover, \topviews{} edits contain more popular subjects (as intended), while \recentlyemerged{} edits have more popular objects.
Fig.~\ref{figure:top_relation_stats} shows the top relations and their frequency in each subset of \ripple{}, demonstrating the diversity of the generated facts.

\subsection{Data Quality}
\label{sec:data_quality}

We conducted a manual analysis to validate that our generation pipeline produces valid test queries. Concretely, we sampled 200 random test queries from \ripple{} and checked the following two requirements: (a) \emph{soundness}: the triplet that represents a given test query should be semantically correct, namely, the entity type of the object should match the relation type and the relation type should match the entity type of the subject. For example, queries such as \emph{``The capital of Hilary Clinton is''} or \emph{``The sibling of Lebron James is Los Angeles''} would have been disqualified. (b) \emph{grammatically correct}: we check that the phrasing of the test query in natural language is grammatical. 

We found that 100\% of the queries were sound (i.e., semantically clear and correct), showing that the data curating process was designed properly.
Furthermore, 98.5\% of the queries were grammatically correct, while the ones which were not contain entity representations in a non-English language. This shows that our templates are general enough to properly fit various entity names.

\section{Experiments}
\label{sec:experiments}

We use \ripple{} to evaluate recent KE methods, and show that despite substantial progress on existing benchmarks, current methods struggle to introduce consistent changes to the model's knowledge after an edit. Moreover, a simple in-context editing baseline that conditions the generation on the edited fact obtains better results, while leaving ample room for improvement for future research.

\begin{figure}[t]
\setlength{\belowcaptionskip}{-8pt}
    \centering \includegraphics[scale=0.48]{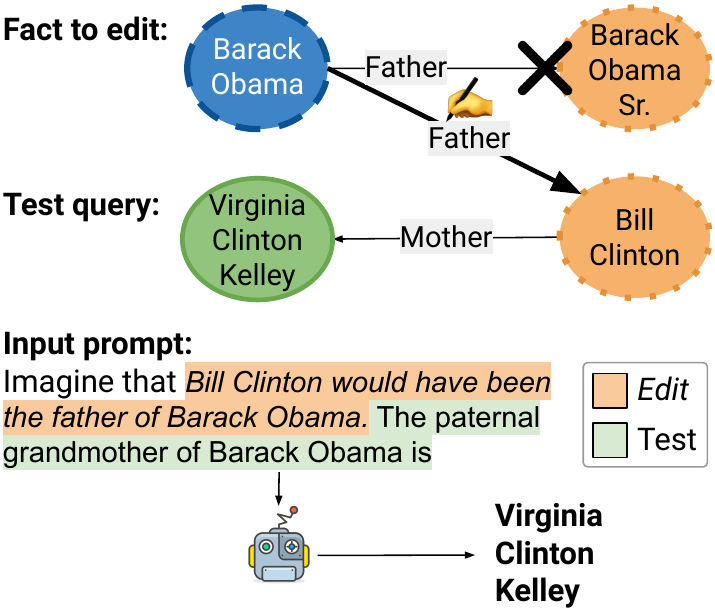}
    \caption{
    An example modification edit from our ICE baseline.
    The color code of the KG is similar to that described in Fig.~\ref{figure:examples}. We prepend the prefix ``\emph{Imagine that}'' to the input prompt, as counterfactuals can contradict knowledge embedded in a model's parameters.  
    }    \label{figure:ice_demonstration}
\end{figure}

\subsection{Evaluation Setting}
\label{subsec:experimental_settings}

\paragraph{Data}
To evaluate how well an editing method handles the ripple effects resulting from editing a given model, the data first needs to be adjusted such that (a) only cases of successful edits are evaluated, and (b) only test queries that the model answered correctly pre-editing are used for evaluation. 
Concretely, for an editing method $\mathcal{F}$ and a model $\mathcal{M}$, an edit request $x: (e,r,o) \rightarrow (e,r,o')$ is included in the evaluation if the following conditions are met when applying $\mathcal{F}$ to $\mathcal{M}$ and $x$: (a) $\mathcal{M}$ successfully generates the original objects for the test queries before applying the edit, and (b) $\mathcal{M}$ successfully generates $o'$ when queried about $e$ and $r$, namely, the edit has successfully been applied. 
For example, we verify that the model can predict the children of $o'$ before asking about $e$'s new siblings.

\paragraph{Editing methods} 
We evaluate three KE methods:
MEND \cite{mitchell2022fast}, ROME \cite{meng2022locating}, and MEMIT \cite{meng2022mass}. MEND trains a network that modifies gradients to produce local edits.
ROME makes rank-one updates to the weights of the Transformer's MLP layers to modify specific factual associations, and MEMIT is an extension of ROME that is adjusted to editing many facts at once.

\paragraph{Baseline} Motivated by the recent success of LMs to learn in-context and follow instructions \cite{NEURIPS2020_1457c0d6, ouyang2022training, liu2023pre},
specifically for knowledge editing \cite{zhong2023mquake, zheng2023edit},
we experiment with an in-context editing (ICE) baseline for factual editing.
Unlike the above methods, it does not introduce changes to the model parameters, but rather generation is conditioned on the new fact. 
Concretely, given an edit $(e, r, o) \rightarrow (e, r, o^*)$ and a test query $q$, we use the following prompt to obtain an answer from the model: \texttt{``Imagine that <$o^*$> would have been <$P_r$>''}, where $P_r$ is a manually-written proposition of $r$, such as \emph{``The mother of <$e$>''} when $r=$ \texttt{Mother} and $e$ is the subject.
An example is depicted in Fig.~\ref{figure:ice_demonstration}.

\begin{table}[t]
\setlength{\belowcaptionskip}{-8pt}
\setlength\tabcolsep{4pt}
    \centering
    \footnotesize
    \begin{tabular}{lllccrr}
    & \multicolumn{2}{c}{\recentlyemerged{}} & \multicolumn{2}{c}{\fakefacts{}} & \multicolumn{2}{c}{\topviews{}} \\ 
    & Edits & Tests & Edits & Tests & Edits & Tests \\ \midrule
        \gptxl{}
         &$853$ &$29\%$ &$689$ &$33\%$ &$722$ &$71\%$  \\ 
         \gptj{}
         &$801$ &$33\%$ &$717$ &$34\%$ &$760$ &$76\%$  \\ 
         \gptneo{} 
         &$989$ &$45\%$ &$801$ &$46\%$ &$828$ &$86\%$ \\ 
         \llama{} 
         &$847$ &$44\%$ &$796$ &$49\%$ &$784$ &$87\%$   \\ 
         \gptt{} 
         &$822$ &$55\%$ &$760$ &$74\%$ &$665$ &$94\%$  \\
         \bottomrule 
    \end{tabular}
    \caption{(a) Number of edits considered in our evaluation (i.e., that have successfully applied), from each subset, averaged over \rome{}, \memit{} and \mend{}, for the models: \gptxl{}, \gptj{}, \gptneo{} and \llama{}, and the ICE baseline for \gptt{}. (b) Portion of queries, on average, that were used in our evaluation.}
\label{table:filtered_tests_portion}
\end{table}

\paragraph{Models} 
We use 4 recent auto-regressive decoder-only LMs of different sizes: GPT-2 XL \cite{radford2019language} with 
1.5B parameters,
GPT-J \cite{chen2021evaluating} with 6B parameters, LLaMA with 7B parameters, \cite{touvron2023llama}, and GPT-NeoX with 20B parameters \cite{black2022gpt}.
In addition, as our baseline does not require access to the model parameters, we also evaluate it on the closed-source model GPT-3 \texttt{text-davinci-003} with 175B parameters \cite{brown2020language}.
However, for the baseline we do not include results for \gpt{} and \gptj{} as the number of testable edits for these models is rather small ($\leq20\%$ for each of the data subsets).

For all model-method combinations, except for \rome{} with \llama{}, we use the official implementation and hyperparameters from \citet{meng2022locating}. 
We adjust ROME to \llama{} by following the authors' method and codebase.
Table~\ref{table:filtered_tests_portion} shows the number of edits and test queries left, for every model, after filtering out non-successful edits and inapplicable test queries (as described above).

\paragraph{Evaluation} Each model-method pair is evaluated separately, on every subset of \ripple{}. For each evaluation criterion, we first compute the average accuracy over the test queries per example, and then average over all the examples. For a given test query, we let the model generate a maximum of 20 tokens. A generation is considered successful if one of the aliases of the target object appears in the text. In cases of multiple gold target objects (as in \forgetfulness{}), we evaluate each target object separately and consider the generation as correct if it matches at least one object.

\begin{table}[t]
\setlength\tabcolsep{3pt}
    \centering
    \footnotesize
    \resizebox{0.999\linewidth}{!}{
    \begin{tabular}{llcccccc|r}
    & & LG & CI & CII & SA & PV & RS & Avg. \\ \midrule
        \multirow{3}{*}{\gptxl{}} 
         & ROME &$20.2$ &$35.6$ &$46.8$ &$86.8$ &$100$ &$55.4$ &$57.5$  \\
         & MEMIT &$21.8$ &$30.3$ &$46.2$ &$92.9$ &$100$ &$56.8$ &$58.0$  \\
         & MEND &$28.9$ &$23.7$ &$20.7$ &$87.1$ &$100$ &$51.9$ &$52.1$  \\ \midrule
         \multirow{2}{*}{\gptj{}} 
         & ROME &$15.2$ &$29.5$ &$50.5$ &$90.3$ &$99.4$ &$60.0$ &$57.5$  \\
         & MEMIT &$18.0$ &$35.0$ &$48.1$ &$88.4$ &$98.6$ &$42.2$ &$55.0$  \\ \midrule
         \multirow{2}{*}{\gptneo{}} 
         & ROME &$27.2$ &$54.3$ &$69.4$ &$98.9$ &$98.4$ &$80.3$ &$71.4$    \\
         & ICE &$48.3$ &$29.0$ &$62.2$ &$100$ &$99.4$ &$80.7$ &$69.9$  \\ \midrule
         \multirow{2}{*}{\llama{}} 
         & ROME &$16.7$ &$47.8$ &$50.0$ &$93.6$ &$97.6$ &$59.3$ &$60.8$  \\
         & ICE &$59.6$ &$74.8$ &$85.0$ &$100$ &$99.5$ &$77.9$ &$82.8$   \\ \midrule
         \gptt{} 
         & ICE &$33.3$ &$100$ &$91.3$ &$100$ &$100$ &$73.1$ &$82.8$  \\
         \bottomrule 
    \end{tabular}
    }
    \caption{Accuracy on the \recentlyemerged{} subset, by \mend{}, \rome{}, \memit{}, and the ICE baseline, on \gptxl{}, \gptj{}, \gptneo{}, \llama{}, and \gptt{}. 
    }
\label{table:recently_emerged_results}
\end{table}

\begin{table}[t]
\setlength{\belowcaptionskip}{-10pt}
\setlength\tabcolsep{3pt}
    \centering
    \footnotesize
    \resizebox{0.999\linewidth}{!}{
    \begin{tabular}{llllccrr|r}
    & & LG & CI & CII & SA & PV & RS & Avg. \\ \midrule
        \multirow{3}{*}{\gptxl{}} 
         & ROME &$53.6$ &$31.6$ &$44.4$ &$94.9$ &$9.9$ &$38.9$ &$45.5$  \\
         & MEMIT &$58.4$ &$30.5$ &$49.8$ &$100$ &$20.0$ &$36.2$ &$49.1$   \\
         & MEND &$62.5$ &$16.7$ &$14.6$ &$91.3$ &$17.7$ &$30.1$ &$38.8$   \\ \midrule
         \multirow{2}{*}{\gptj{}} 
         & ROME &$53.8$ &$40.8$ &$49.9$ &$93.8$ &$15.2$ &$39.4$ &$48.8$   \\
         & MEMIT &$53.0$ &$35.7$ &$48.2$ &$95.6$ &$18.2$ &$39.9$ &$48.4$   \\ \midrule
         \multirow{2}{*}{\gptneo{}} 
         & ROME &$61.6$ &$49.4$ &$57.1$ &$100$ &$30.8$ &$50.7$ &$58.3$  \\
         & ICE &$78.6$ &$90.0$ &$55.6$ &$100$ &$100$ &$61.9$ &$81.0$    \\ \midrule
         \multirow{2}{*}{\llama{}}
         & ROME &$54.3$ &$35.5$ &$49.5$ &$96.0$ &$17.8$ &$38.9$ &$48.7$ \\
         & ICE &$71.1$ &$73.8$ &$80.3$ &$100$ &$100$ &$69.6$ &$82.5$\\  \midrule
         \gptt{} 
         & ICE &$69.0$ &$83.3$ &$89.7$ &$100$ &$100$ &$100$ &$90.3$  \\
         \bottomrule
    \end{tabular}
    }
    \caption{Accuracy on the \fakefacts{} subset, by \mend{}, \rome{}, \memit{}, and the ICE baseline, on \gptxl{}, \gptj{}, \gptneo{}, \llama{}, and \gptt{}.}
\label{table:fake_facts_results}
\end{table}

\begin{table}[t]
\setlength\tabcolsep{3pt}
    \centering
    \footnotesize
    \resizebox{0.999\linewidth}{!}{
    \begin{tabular}{llllccrr|r}
    & & LG & CI & CII & SA & PV & RS & Avg. \\ \midrule
        \multirow{3}{*}{\gptxl{}} 
         & ROME &$5.7$ &$46.4$ &$21.8$ &$100$ &$100$ &$18.5$ &$48.7$  \\
         & MEMIT &$6.7$ &$45.2$ &$21.2$ &$100$ &$100$ &$24.3$ &$49.6$   \\
         & MEND &$25.9$ &$10.7$ &$5.4$ &$100$ &$100$ &$21.2$ &$43.9$   \\ \midrule
         \multirow{2}{*}{\gptj{}} 
         & ROME &$5.5$ &$44.1$ &$21.0$ &$98.6$ &$99.0$ &$22.3$ &$48.4$   \\
         & MEMIT &$7.0$ &$45.9$ &$23.7$ &$100$ &$100$ &$24.8$ &$50.2$   \\ \midrule
         \multirow{2}{*}{\gptneo{}} 
         & ROME &$36.4$ &$29.4$ &$41.6$ &$100$ &$100$ &$50.8$ &$59.7$  \\
         & ICE &$37.5$ &$92.4$ &$40.1$ &$100$ &$100$ &$74.4$ &$74.1$    \\ \midrule
         \multirow{2}{*}{\llama{}} 
         & ROME &$22.0$ &$37.4$ &$16.2$ &$100$ &$100$ &$20.6$ &$49.4$ \\
         & ICE &$57.2$ &$85.1$ &$67.6$ &$100$ &$100$ &$78.0$ &$81.3$   \\ \midrule
         \gptt{} 
         & ICE &$31.0$ &$86.1$ &$65.6$ &$100$ &$100$ &$83.8$ &$77.7$  \\
         \bottomrule
    \end{tabular}
    }
    \caption{Accuracy on the \topviews{} subset, by \mend{}, \rome{}, \memit{}, and the ICE baseline, on \gptxl{}, \gptj{}, \gptneo{}, \llama{}, and \gptt{}.}
\label{table:top_views_results}
\end{table}

\subsection{Results}
Tables~\ref{table:recently_emerged_results},~\ref{table:fake_facts_results},~\ref{table:top_views_results} show the evaluation results on the \recentlyemerged{}, \fakefacts{}, and \topviews{} subsets, respectively.
Considering the average scores across all subsets, we observe that existing editing methods struggle to handle the ripple effect induced by editing facts, with low average accuracy of $38-66$ across all models.
This suggests that, while KE methods demonstrate high capability in making local updates to the model's knowledge, these changes are mostly applied at a surface-level without propagating to other related facts.
Moreover, we observe that our ICE baseline obtains the best overall results. Specifically, it outperforms \rome{} by more than 10 points for \gptneo{} and 29 points for \llama{}, on average across subsets. Although \gptt{} with ICE performs best on average, the 7B \llama{} is highly competitive, performing better or similarly on the \recentlyemerged{} and \topviews{} subsets.

Next, comparing results across evaluation criteria shows that some ripple effects are handled better than others. For example, while \aliasing{} accuracy is consistently high ($\geq86.8$ across all settings), the accuracy on other criteria is generally lower and varies greatly between models, methods, and edits (e.g., \logicalgeneralization{} accuracy for \rome{} on \gptj{} is $53.8$ on the \fakefacts{} subset, compared to only $5.5$ on the \topviews{} subset).

\begin{figure}[t]
    \centering
    \includegraphics[scale=0.6]{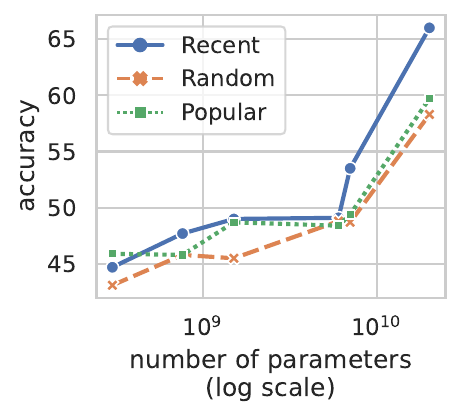}
    \caption{Accuracy averaged over evaluation criteria of \rome{}, as a function of the model's number of parameters, for the following models: \textsc{GPT2-M}, \textsc{GPT2-L}, \textsc{GPT2-XL}, \gptj{}, \llama{}, and \gptneo{}.} \label{figure:acc_as_a_function_of_size}
\end{figure}

\paragraph{Results across model size}
We analyze how editing performance on \ripple{} is influenced by the model size. To this end, we further evaluate \rome{} on smaller versions of \gpt{} -- with 345M (GPT2-M) and 762M (GPT2-L) parameters, and plot the average accuracy over the three subsets as a function of model size.
Fig.~\ref{figure:acc_as_a_function_of_size} presents the results, showing that editing performance increases with model size, with \rome{} obtaining substantially higher accuracy when applied to larger models. 
Nevertheless, our results (Tables~\ref{table:recently_emerged_results},~\ref{table:fake_facts_results},~\ref{table:top_views_results}) show that when using ICE, the 7B \llama{} is competitive with the much larger \gptt{}, suggesting that simply scaling the model size may not be sufficient to fix the drawbacks of current editing methods. 

\paragraph{Results across methods}

\begin{table}[t]
\footnotesize
\begin{center}
\begin{tabular}{lccc}
 & \mend{} & \rome{} & \memit{} \\ [0.1cm]
\toprule
\relationspecificity{}          & $34.4$    &$37.6$   &$39.1$  \\
\logicalgeneralization{}    & $39.1$    &$26.5$   &$29.0$  \\
\compositionality{}            & $17.0$    &$37.9$   &$35.3$   \\ 
\forwardcompositionality{}    & $13.6$    &$37.7$   &$39.1$    \\
\bottomrule
\end{tabular}
\end{center}
\caption{Accuracy of \mend{}, \rome{} and \memit{}, using \gptxl{}, averaged over the three \ripple{} splits - \recentlyemerged{}, \fakefacts{} and \topviews{}.}
\label{table:res_across_methods}
\end{table}

Table~\ref{table:res_across_methods} shows the accuracy of \mend{}, \rome{} and \memit{}, on \gptxl{} across our evaluation criteria, averaged over the three subsets. Interestingly, \mend{} outperforms \rome{} and \memit{} in \logicalgeneralization{}, but is worse in \compositionality{} and \forwardcompositionality{}, suggesting that different methods might better capture different types of ripple effects.

\begin{figure}[t]
    \centering
    \includegraphics[scale=0.5]{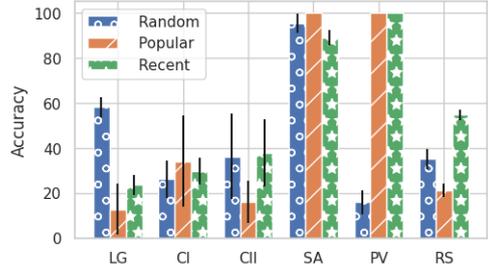}
    \caption{The average accuracy of \gpt{} on different evaluation criteria in \ripple{}. Results are averaged over editing methods (\rome{}, \memit{} and \mend{}); error bars indicate standard deviation.} \label{figure:test_scores_per_split}
\end{figure}

\paragraph{Results across data splits}
The subsets of \ripple{} differ in whether edited facts are counterfeit
or real, and in the popularity of the edited entities. These differences allow us to control for the edit severity, as popular entities are expected to introduce larger ripple effects (see \S\ref{sec:rethinking}).
In Fig.~\ref{figure:test_scores_per_split}, we show the accuracy on each subset and evaluation criterion, averaged over the different editing methods.
Comparing \fakefacts{} and \topviews{}, that differ in the popularity of the edited entities, we see that while \logicalgeneralization{} accuracy is substantially higher for \fakefacts{}, \forgetfulness{} accuracy is higher for \topviews{}. This suggests that, although retaining correct knowledge is easier for popular entities, modifying other facts that logically follow from an edit is harder for popular entities, which could be explained by the severity of these edits (i.e. the high number of facts that are semantically related to them).

\begin{table}[t]
\setlength{\belowcaptionskip}{-10pt}
\setlength\tabcolsep{3pt}
    \centering
    \footnotesize
    \begin{tabular}{llccc}
    & & No effect & Abstaining & Noise \\ \midrule
        \multirow{2}{*}{\gptxl{}} 
         & ROME & 27\% & 31\% & 42\%  \\
         & ICE & 32\% & 27\% & 41\%  \\ \midrule
         \multirow{2}{*}{\gptneo{}} 
         & ROME & 24\% & 40\% & 36\%  \\
         & ICE & 10\% & 65\% & 25\%     \\ \midrule
         \multirow{2}{*}{\llama{}}
         & ROME & 20.5\% & 45\% & 34.5\% \\
         & ICE & 11\% & 71\% & 18\%  \\
         \bottomrule
    \end{tabular}
    \caption{Error type distribution on 200 failures of \rome{} and ICE, on \gptxl{}, \gptneo{}, and \llama{}.}
\label{table:error_analysis}
\end{table}

\subsection{Error Analysis}
\paragraph{\rome{} versus ICE}
We qualitatively analyze the effect induced by KE methods to the model's knowledge. To this end, for each of \rome{} and our ICE baseline and each of the models \gpt{}, \gptneo{}, and \llama{}, we sample 200 test queries from \ripple{} on which the model fails post-editing. We then label these failures using three categories: (a) \textit{no effect}, for cases when the model predicts the original object, i.e. the edit introduced no ripple effect, (b) \textit{abstaining}, when the model abstains from answering by generating text like \textit{``unknown''} or \textit{``a mystery''}, and (c) \textit{noise}, when the model generates an incorrect object or unrelated text.
Table~\ref{table:error_analysis} presents the results, showing that in most cases ($\geq68\%$ across all settings) factual editing introduces erroneous changes to the model's knowledge rather than making no change. Interestingly, for both \gptneo{} and \llama{}, where editing performance is better than \gpt{}, \rome{} introduces more incorrect changes while ICE causes the model to abstain from answering.

\paragraph{\gptt{} versus \llama{} using ICE}
We further looked into the performance on the LG tests, where applying ICE to \gptt{} is notably inferior to ICE on \llama{} (see Tables~\ref{table:recently_emerged_results},~\ref{table:fake_facts_results},~\ref{table:top_views_results}). Specifically, we collected responses from each of the models to 100 random LG queries, and analyzed them using the same categories as described above. 
We observed that \gptt{} abstains from answering the query much more often than \llama{} (49\% of the cases for \gptt{} compared to only 28\% in \llama{}), which could explain the lower performance of ICE on \gptt{} on these queries.

\section{Conclusion and Discussion}

We introduce the notion of ripple effects in knowledge editing, suggesting that editing a particular fact implies further updates of related facts. We additionally propose evaluation criteria for ripple effects and create \ripple{}, a diagnostic benchmark designed to evaluate how well KE methods handle the ripple effects of various edits. 
We evaluate prominent KE methods and show that they often fail to introduce consistent edits that capture the ripple effects of an edit, suggesting that future development of KE methods should consider those effects more carefully. Last, we show that a simple in-context editing method achieves the best results on \ripple{}, highlighting the potential of such editing approaches.

Notably, our benchmark covers a small fraction of all possible ripple-edits. For example, one could consider ripple effects that involve more than two hops, and explore the graph structure of different edits.
In addition, while we focus on ripple effects of single edits, future work can consider the effect of editing multiple facts in a single batch. Finally, it would be interesting to consider cases where models succeed in capturing ripple-edits, and analyze how these are implemented mechanistically in the transformer architecture \cite{geva2023dissecting}.

\paragraph{Limitations}
Our data generation pipeline relies on information from an existing knowledge-base (\wikidata{} in our case), which could be incomplete or outdated. While \ripple{} does not aim to cover all the possible ripple-edits in \wikidata{}, these concerns might be a major issue when seeking a comprehensive evaluation or considering domain-specific knowledge-bases, which often tend to be incomplete. A possible solution to explore in that case is to use LMs internal knowledge instead of an external knowledge-base \cite{cohen-etal-2023-crawling}.

With \ripple{} focusing on the ripple effect of edits, it does not include tests, such as paraphrasing of the edit and subject specificity, that evaluate the edit itself and are covered by existing benchmarks (e.g. CounterFact). In addition, it does not verify that many other facts that are distantly related to the edit, i.e., triplets that are not included in the close neighbourhood of the edit, were retained post-editing. For example, we expect that editing the capital of France would not affect the population of Poland, yet this is not explicitly checked. We note that building such an evaluation is hard, since there are many facts to consider and it is unclear how to determine automatically which triplets should and should not be affected by a certain edit.

\iftaclpubformat

\else
\fi

\section*{Acknowledgments}
We thank Maor Ivgi and Gal Elidan for valuable feedback and constructive suggestions.
This work is supported in part by the Israeli Science Foundation.

\bibliography{tacl2021}
\bibliographystyle{acl_natbib}

\iftaclpubformat

\onecolumn

\appendix






  
\fi

\end{document}